\documentclass{article}
\pdfoutput=1
\usepackage[english]{babel} 
\usepackage[T1]{fontenc}
\usepackage[utf8x]{inputenc}
\usepackage{amsfonts}
\usepackage{amsmath}
\usepackage{hyperref}
\usepackage{graphicx}
\usepackage{subcaption}
\usepackage{enumitem}   
\usepackage{float}
\usepackage{afterpage}
\usepackage[export]{adjustbox}
\usepackage{authblk}

\newcommand{\uniboaffil}{Department of Computer Science and Engineering (DISI),$\;\;\;\;$ {\it Alma Mater Studiorum} Universit{\`a} di Bologna, Italy}
\newcommand{\isbaffil}{Institute for Systems Biology, Seattle, USA}
\newcommand{\ecltaffil}{European Centre for Living Technology, Venice, Italy}

\title{Online adaptation in robots as biological development provides phenotypic plasticity}

\author[1]{Michele~Braccini}
\author[1,2]{Andrea~Roli}
\author[3]{Stuart~Kauffman}

\affil[1]{\uniboaffil}
\affil[2]{\ecltaffil}
\affil[3]{\isbaffil}

\date{}

\begin{document}

\maketitle

\begin{abstract}
The ability of responding to environmental stimuli with appropriate actions is a property shared by all living organisms, and it is also sought in the design of robotic systems. Phenotypic plasticity provides a way for achieving this property as it characterises those organisms that, from one genotype, can express different phenotypes in response to different environments, without involving genetic modifications. In this work we study phenotypic plasticity in robots that are equipped with online sensor adaptation. We show that Boolean network controlled robots can attain navigation with collision avoidance by adapting the coupling between proximity sensors and their controlling network without changing its structure. In other terms, these robots, while being characterised by one genotype (i.e. the network) can express a phenotype among many that is suited for the specific environment. We also show that the dynamical regime that makes it possible to attain the best overall performance is the critical one, bringing further evidence to the hypothesis that natural and artificial systems capable of optimally balancing robustness and adaptivity are critical. 
\end{abstract}

\section{Introduction}

Since Darwin's and Mendel's works on the theory of evolution by natural selection~\cite{darwin2004origin} and on the laws of biological inheritance~\cite{mendel1866versuche} respectively, evolutionary theories have focused mainly on the role of selection acting on randomly-generated genetic material in the origination of \emph{phenotypic diversification}---and finally \emph{speciation}.
This way of thinking---summarised by the so-called ``Neo-Darwinian Synthesis'' or ``Modern Synthesis''---fostered the idea that the responsibility for the generation and the subsequent establishment of ``evolutionary novelty'' was prerogative of genetic material.
So, the differential survival and reproduction success of biological organisms has been ascribed to genotype.

Parallel to the development of ``Modern Synthesis'', Mayr~\cite{Mayr2091}  argued that: ``[...] it is the
phenotype which is the part of the individual that is \emph{visible}
to selection.''
Mayr's argument, together with Waddington's work on the \emph{Epigenetic Landscape}~\cite{waddington1942epigenotype}, has paved the way for the formulation of a theoretical framework according to which the phenotype---and not the genotype---,the environment~\footnote{With the term \emph{environment} here we refer---without loosing generality---to all external perturbations from which a subject can be influenced, some of these could be the external world itself, organisms of the same or other species, etc.} and above all the development process play a primary role in the origin of the novelty from an evolutionary point of view.

The epigenetic landscape metaphor, which finds a formal basis in dynamical systems theory, stresses the concept for which there is no trivial deterministic mapping between genotype and phenotype~\cite{huang2012molecular}.
It is the dynamics of the complex network of interaction among genes, and between genes and the environment, which will determine the stable expression patterns and so ultimately will affect the phenotype determination.
Therefore, the ensemble of dynamics than can be generated by the genes composing the organism's genetic code represents a source of diversification that can explain the birth of new phenotypes and, consequently, their affirmation on the evolutionary scale. It is important here to emphasise the role of the environment in constraining and shaping these actual dynamics. 
In biology, the capacity of a genotype to produce different phenotypes depending on the environment in which it is located is defined as \emph{phenotypic plasticity}~\cite{PFENNIG2010459,kelly2011phenotypic}, \emph{developmental plasticity} if differences emerges during development~\cite{fusco-minelli-2010,gilbert2016developmental}.
The specific dynamics that shapes an organism's phenotype during its development is indeed the response to various influences, among which we inevitably find the external environment, other organisms and noise~\cite{longo_how_2018}.
More in general, these external agents influence the process of regulation so that they might destabilise reached (meta)stable patterns of gene expression and induce a network dynamics reconfiguration, able to accommodate and possibly give appropriate responses to the new state of the external environment. In other words, they stimulate the process of construction of a new internal model of the external world.
Biologists call this process \emph{developmental recombination}; in the works~\cite{West-Eberhard6543,PFENNIG2010459} reasons and evidences why this process is held responsible for the origin of differences between species are presented.
Noteworthy is the hypothesis for which the phenotypic plasticity would be able to allow the crossing of the valleys present in the fitness landscape, a crossing that would be precluded to evolution-by-mutations as the valleys' phenotypes would be selectively disadvantageous.
Therefore, even if mutations (random or not) contribute to the creation of diversification, by modifying the gene regulatory networks topology and therefore the constraints imposed on its dynamics, they are not necessary condition for the phenotypic plasticity and, in the light of the previous discussion, assume a role of supporting actors.
They are, however, implicated in the \emph{genetic accommodation} process~\cite{West-Eberhard6543}, that is the process following the selection of the phenotypic variant with a genetic component; or when a reorganisation of the genotype allows individuals of subsequent generations to reach the same phenotype at a lower cost~\cite{bateson_gluckman_2011}, in terms of time, resources, etc.

From a cybernetics point of view~\cite{wiener1948cybernetics,ashby-cybernetics}, this capability of producing an internal model of the external world, which is provided by phenotypic plasticity, is of great importance. In abstract terms, this process makes it possible for an organism to  compress the wealth of information coming from the external world into an internal representation that values only the pieces of information relevant for the organism's survival; on the basis of this internal model the organism acts so as to achieve its tasks, \textit{in primis} to attain homeostasis, i.e. maintaining its \textit{essential variables} within physiological ranges~\cite{design-for-a-brain}. We can then state that phenotypic plasticity not only is a vital property for living organisms, but it may also be of great value for artificial ones.
Therefore, a fundamental question arises as to what are the \emph{generic properties} that allow organisms to exhibit the phenotypic plasticity observed in the process of organisms development and so attain an effective level of adaptivity.
If these properties are found, on the one hand they may provide us insights about the mechanisms underlying the adaptive behaviours of organisms during their development process; 
on the other hand, given the reported relevance that the development process may have on evolutionary-scale changes~\cite{arthur_2004}, they can be the key to understand the onset of the differences, and at the same time the common traits, between the species~\footnote{Evolutionary Developmental Biology (evo-devo) was born around the end of the twentieth century with the intention of answering these and other related questions.
In particular, it focuses on the role of the developmental process, and the effects of its alteration, on evolutionary changes~\cite{Hall2012,WallaceEvoDevo}}.
An approach based on generic properties provides an alternative to the comparative studies between different species, which although have led to great results (see above all the discoveries of \emph{homeobox} and \emph{Pax6} gene, ~\cite{Gehring2007,WallaceEvoDevo,Hall2012,Xu383}) have the limitation of being highly costly and not being easily generalisable.
In addition, general properties supporting phenotypic plasticity may provide an effective design principle for artificial systems capable to adapt.

To this aim we believe it is necessary to start from the known and most relevant properties of the organisms and check whether they can also provide plausible hypotheses for the construction of general principles that can first explain the phenotypic plasticity and then hopefully can bring us to link development and evolution.
We believe that one of these principles can be found in \textit{criticality}.
A long-standing conjecture in complex system science---the \emph{criticality hypothesis}---emphasises the optimal balance between robustness and adaptiveness of those systems that are in a dynamical regime between order and chaos~\cite{Kauf93,Kau1996}.\footnote{A recent account of dynamical criticality can be found in ~\cite{roli2018dynamical,munoz2018colloquium}.}
Theoretical studies on properties of such systems and a bunch of empirical evidences led to a reshape of this conjecture into ``life exists at the edge of chaos''~\cite{Lan1990,Pac1988}, or in the field of information processing into
``computation at the edge of chaos''~\cite{CruYou1990,Pro2013}.
Among the most remarkable experimental studies that have brought evidence for the criticality hypothesis, we focus our attention on those that belong to the field of biology, since we are interested in biological development and evolution.

In many papers, it emerges that biological cells---or more precisely their gene networks---operate in the critical dynamic regime.
This has been repeatedly corroborated by different authors and with the help of different techniques, models and working hypotheses.
The comparison of time sequences of microarray gene-expression data against data generated by random Boolean networks (RBNs) models~\cite{kauffman69} led Shmulevich and others~\cite{shmulevich2005eukaryotic}, to conclude that genetic regulatory network of HeLa cells---an eukaryotic cancer cell line---operate in the ordered or critical regime, but not in chaotic one.
In~\cite{DanielsWalkerCriticality}, by exploiting the CellCollective~\cite{helikar2012cell} database of Boolean models of real regulatory networks, the authors showed that by using the \emph{sensitivities} measure~\cite{Sensitivities} all the network taken in exam were critical, or near critical.
Further, Serra and Villani showed that Boolean networks that best fit the knock-out avalanches in the yeast \textit{S.~Cerevisiae} are ordered, but very close to the critical boundary.

Many noteworthy papers~\cite{beggsCriticality,BeggsBrain,ChialvoNeural} that focus on analyses of the brain, mainly making use of models, brought evidences supporting that also the brain works in critical condition.
As an example, evidence concerning models of C. Elegans' nervous system activities in free locomotion condition shows that its brain functioning has some signature of criticality~\cite{cElegansIzquierdo}. 

At the same time, there is evidence suggesting that organisms that operate in critical condition are the most advantaged, evolutionarily speaking.
Aldana et al. in~\cite{AldBalKauRes2007} pointed out that well-known model of biological genetic networks, RBNs (formally introduced in the following), in critical regime showed the properties of robustness and, in particular, \emph{evolvability}, at the same time.
More in detail, they introduced network \emph{mutations} (see experimental details in the original paper) and assessed the degree to which the original attractors (those exhibited before mutations) are retained and, simultaneously, the capacity to give rise to new attractors.
Although both ordered and critical networks have proven capable of retaining the original attractors with high probability, critical networks were those with the greatest tendency to produce new attractors, and so \emph{evolvability}.  
Torres-Sosa and others~\cite{TorresSosa} have reached the same conclusion by following a slightly different path.
By modelling natural selection acting on Boolean networks as an evolutionary algorithm with mutation and gene duplication operators, they showed that dynamical criticality emerges as a consequence of the attractor landscape evolvability property at the evolutionary level.

A remarkable property of critical systems is that they can reliably respond to the inputs while being capable to react with a wide repertoire of possible actions~\cite{kauffman2000investigations}: this functionality is essential for organisms that must select, filter and compress the information coming from the environment that is relevant for their life.

At this point, we wonder if criticality can foster the emergence of phenotypic plasticity and, if so if this can be at the same time the responsible of the establishment of robustness and adaptivity properties, characteristic of organisms produced by both phylogenesis and ontogenesis.

In this paper we begin tackle these fundamental, open questions; so we start investigating if dynamical criticality can favour phenotypic plasticity.
Indeed, if this were to be the case, not only we could bring further evidence to the ``criticality hypothesis'', but we could start to shed light on the relationship between phenotypic plasticity, criticality and evolution.

\section{Creation of novelty in robotics}

In this work, we make use of robotic agents as a proxy to start investigating the questions raised in the previous section.

The robotics literature is full of examples in which techniques inspired by the natural world are used to allow robots---or swarms of robots---to perform complex tasks~\cite{bonabeau1999swarm,nolfi-evorobot-book,braccini2017applications}.
Dually, we can think of employing artificial agents for representing or mimicking natural dynamics and therefore, to investigate issues and open questions otherwise impossible to study.
Indeed, robots development and design costs are very low (especially considering the possibilities offered by  simulation) and therefore, ideal for these analyses.
Although this artificial approach has intrinsic limits---their results are not easily generalisable \emph{as are} to the natural counterpart---they can provide us with some new clues, hypotheses or different perspectives that could lead to the formation of new models or theories, besides suggesting specific experiments to be undertaken.

Artificial devices have already proven to be able to give rise to the emergence of diversity and the creation of novelty.
In this regard, Gordon Pask in the 1950s conducted a remarkable experiment involving truly evolvable hardware~\cite{pask1958physical,pask1960natural,cariani1993evolve}.
Pask builded an electrochemical device with emerging sensory abilities.
In particular, the experimental structure he created---composed of electrodes immersed in a ferrous sulphate solution---was able to evolve from scratch the ability to recognise sounds or magnetic fields.
In other words, the assembly~\footnote{It can be considered an example of evolutionary robotic device~\cite{cariani1992some}.} developed its own sensors from scratch and therefore its own \emph{relevance criteria} from the outside world.

Inspired by Pask's works, Peter Cariani proposes a classification of the kinds of adaptive behaviours attainable by physical devices~\cite{cariani1992some,Cariani2008EmergenceAC}.
He calls \emph{nonadaptive robotic devices} those devices that are not able to modify their internal structure based on their past experiences.
\emph{Adaptive computational devices} is instead the category for devices that can change their computational module, if advantageous. 
They can improve only the mapping between their fixed input and output though. 
With the term \emph{structurally adaptive devices} he refers to devices capable of constructing new hardware, sensors and effectors for themselves.
As Cariani points out, this is analogous to the biological evolution of organs.
Through the building of sensors and effectors, and so with the freedom to gather and manipulate the kind of information needed to perform a given task, robotic agents can create new semantic states: \emph{relevance criteria} different from those that the designer may have imposed on the robot, initially.
Cariani identifies this last category as a necessary condition for the construction of agents with \emph{epistemic autonomy}: in this condition, the agent is completely autonomous also as regards the creation of new perspectives or point of views of itself, of the world in which is immersed and the relationship among them.

Although they follow a more abstract approach than Pask's pioneering work, attempts to evolve sensors into robotic agents are present in the literature.
For example, in the work~\cite{mark_framework_1998}, both the number and size of sensors are evolved in a simulated environment, finding a preliminary correlation between the number of sensors (in this case artificial eyes) and complexity of the task achieved.
A different---and in some ways more general---perspective is pursued by those approaches that deals with the coevolution of agents' body and brain~\cite{lund_evolving_1997,Dellaert_1996,bongard_evolving_2003}.
Although many of these are inspired by cellular processes and by biological development mainly, these are \emph{offline} approaches.
There are works that tries to apply some elements of evolution in robotics in an online setting~\cite{bredeche2018embodied}, as well as a kind of online epigenetic adaptation~\cite{brawer2017epigenetic}.\footnote{Of course, here we do not consider here all the works concerning generic online adaptation of some parameters of the robotic system.}

The relevance of development in the designing of artificial agents has so far been underestimated; only recently its prospective role and the properties that can derive from it (phenotypic plasticity above all) in the robotic field are emerging~\cite{hunt_phenotypic_2020,jin2011morphogenetic}.
The development process in the designing of artificial entities, by analogy with the biological counterpart, can be represented by an \emph{online adaptation process}~\footnote{The one we propose in the following sections is just a possible example of it.}.
According to the above reported biological evidence and preliminary experiments conducted in artificial contexts, this mechanism has the potential to bring out novelties in the behaviour of artificial agents.
In this context we define \emph{novelty} as the behaviour or outcome reached by the artificial agents that would not have been contemplated by the model the designer has about the robot.
Indeed, the robots actual perceptual experiences and real contingencies give the agent the opportunity to overcome and override the initial designer's constraints and biases.

To conclude, development-inspired (online) approaches are viable alternatives, or complementary tools, for techniques of automatic design of robotic agents.

\section{Proof of Concept}
In light of previous discussions, we \emph{(i)} propose an \emph{online} adaptive mechanism capable of giving rise to the observable phenotypic plasticity property typical of biological organisms without requiring mutations; \emph{(ii)} start analysing the conditions under which robotic agents---using the mechanism referred to in the previous point---obtain a measurable advantage.

Firstly, we briefly introduce the Boolean networks model, since it represents the actual substrate on which our mechanism is based.
Boolean networks (BNs) have been introduced by Kauffman~\cite{kauffman69} as an abstract model of gene regulatory networks.
They are discrete-state and discrete-time non-linear dynamical systems capable of complex dynamics.
Formally, a BN is represented by a directed graph with $n$ nodes each having a variable number $k$ of incoming nodes, a Boolean variable $x_i$, $i = 1,...,n$ and a Boolean function $f_i = (x_{i_1},...,x_{i_k})$.
They have received much attention not only for their capacity to reproduce important properties of biological cells~\cite{shmulevich2005eukaryotic,roli2018dynamical,serra2006,helikar2012cell,DanielsWalkerCriticality} but also for their capacity to express rich dynamics combined with their relatively compact description, characteristics that make them appealing also for artificial applications.
For this reason, the so-called Boolean network robotics takes advantage of them by employing BN instances as control software in robots.
Some examples of the remarkable results that could be obtained through their employment are reported in the following works~\cite{roli2012preliminary,roli-aiia2015,RoliAttractorLandscape}.

The approach we propose---which is grounded on the BN-robotics---consists of using a Boolean network as a robot program.
Its dynamics, in a way similar to that performed by gene regulatory networks in biological cells~\cite{braccini2017applications}, determines the behaviour of the robot and ultimately its final phenotype.
The word \emph{phenotype} is used in this context with a more generic meaning than its biological counterpart: regardless of the specific physical characteristics of the robot, it identifies the overall observable behaviour achieved by the artificial agent.

As illustrated in~\cite{roli2011design} the first step to take when designing a robot control software based on Boolean networks is to choose the coupling between the nodes and the robot actuators and sensors.
Usually, this mapping is chosen at design-time and stay the same during all the design, simulation and, possibly, real-world applications phases.
The mapping itself can be subject to optimisation during the design phase, but once reached the desired/optimal level of robot performance---according to a defined fitness function---it will not undergo any variation.
These approaches are referred to as offline design methods.

With the intention of conferring the property of phenotypic plasticity observed in the development phase of biological organisms (see \emph{(i)}), we propose a novel \emph{online} adaptive mechanism for the design of control software for robots
~\footnote{In the present discussion, we will refer only to robots with fixed morphology, although this mechanism finds natural application in self-assembling robots. They indeed can build their own sensors and really capture their relevance criteria.}.
The BN chosen as control software for the robotic agent is generated once and remains unchanged during all the robot's life.
What distinguishes our approach from past ones is the fact that what changes is the coupling between the BN nodes and the sensors of the agents~\footnote{Although this mechanism is abstract enough to be able to contemplate the variations of both sensors and actuators, in this discussion, we will consider varying only the former.}.
The coupling changes are not externally imposed by the designer: the robot determines which couplings are more suitable to it by continually interacting with the environment in which it is located.

The task chosen for our proof of concept is that of navigation with obstacle avoidance.
The robot, equipped with proximity sensors, must, therefore, try to move in its environment, represented by an arena and at the same time avoid colliding with the obstacles present in it.
This problem can be formally considered as a dynamic classification problem.
Indeed, the artificial agent is faced with a problem of classification of time series which are not independent of the agent's behaviour since they are conditioned by the dynamics belonging to its \emph{sensorimotor loop}~\cite{lungarella2001robots}.

\begin{figure}[h!]
  \centering
    \includegraphics[width=.9\textwidth]{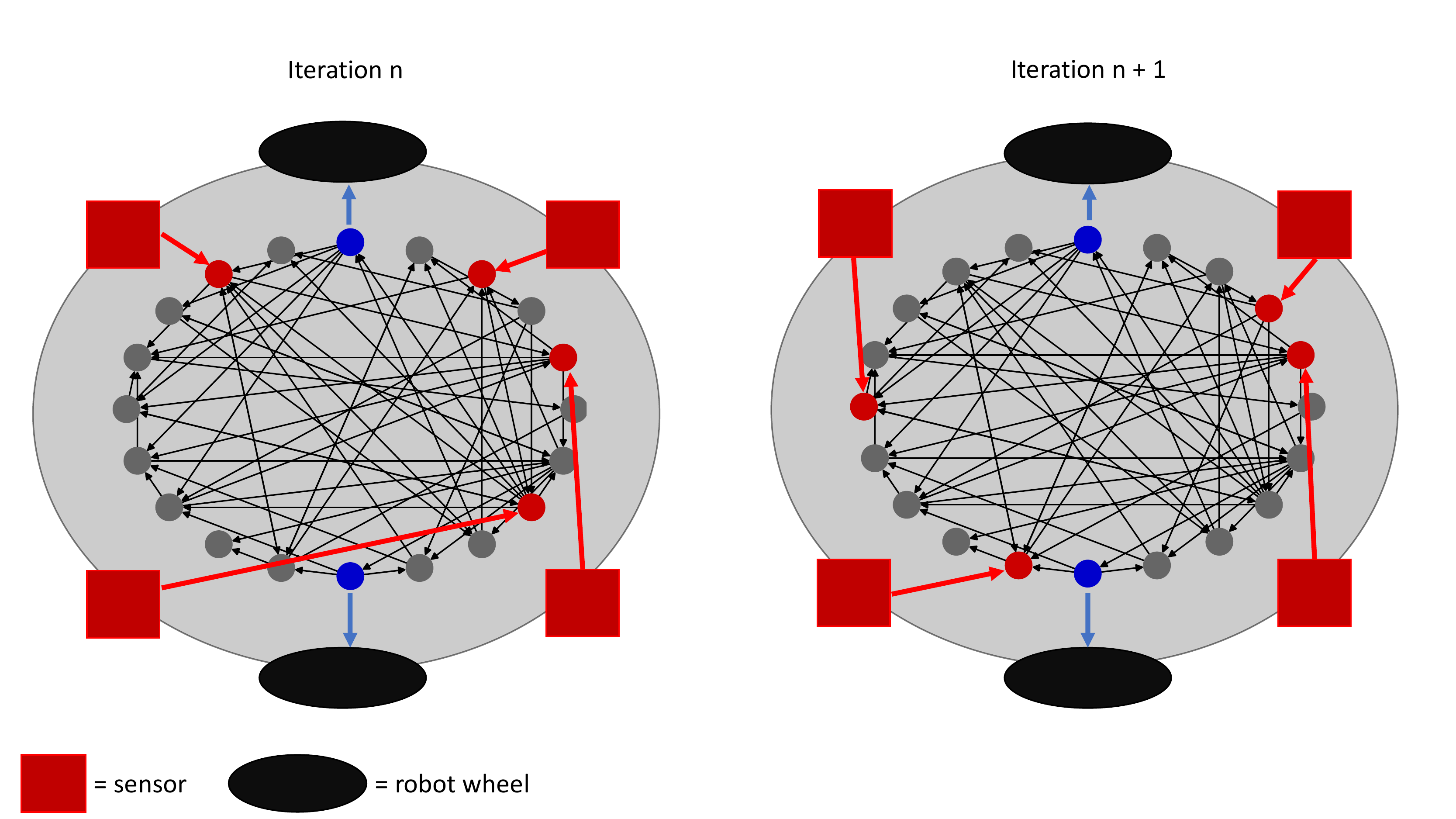}
  \caption{Schematic representation of two consecutive steps of the proposed online adaptive mechanism. The topology of the network, the number of sensors, as well as the nodes coupled with them are only used for example purposes and do not reflect the experimental conditions used in our experiments.}
  \label{fig:img_adaptive_sensors_image_pp}
\end{figure}

Through a designer defined objective function, we provide a figure of merit assessing the degree of adaptation attained by the robot. This function will act as selective pressure and guide the robot adaptation process. It should be considered as an abstraction of the rewarding mechanisms (both intrinsic and extrinsic) that characterise adaptation in natural and artificial systems.
In Figure~\ref{fig:img_adaptive_sensors_image_pp}, we see a schematic representation of two consecutive steps of the process.

Albeit in a more abstract form, this mechanism takes inspiration from the Pask's evolvable and self-organising device.
Here, the space of possibilities among which the robot can choose it's not open-ended, like the one used in Pask's experiment, but it is limited by the possible set of dynamics of the Boolean network, the number of sensors and the coupling combinations between the two.
Simultaneously, it can be considered an artificial counterpart of the adaptive behaviour without mutations present in the development phases of biological organisms.
Indeed, the robot exploits the feedbacks it receives from the environment and consequently tries to re-organise the \emph{raw genetic material} it owns.
In doing so, it does not modify the Boolean network functions or topology but it will use the intrinsic BN's information processing capabilities.

In addition, our adaptive mechanism resembles a step that takes place in the biological phenomenon of neuroplasticity or brain plasticity~\cite{neuroplasticity}.
In neuroplasticity the creation of synaptic connections and changes to neurones occur mainly during the development phase.
However, the process of refinement of the neural network that starts at birth plays a crucial role.
This last occurs as a function of the environmental stimuli and feedbacks the individual receives after his activities and interactions with it. 
The different ensembles of cognitive skills, sensorimotor coordination and, in general, all processes influenced by the brain's activities which an individual develops through his experience represent the sets of possible phenotypes.
At a very high level of abstraction, our ``scrambling phase''---that which change coupling among BN nodes and robot sensors phase---acts as the activity-driven refinement mechanism found in the child's brain.

A further investigative idea behind this experiment and expressed in point \emph{(ii)} is to start finding out what general principles govern the best performing robotic agents, and therefore they promote and at the same time take advantage of the phenotypic plasticity characteristic of our adaptive mechanism.
Fortunately, the literature relating to Boolean networks provides us with a wide range of both theoretical and experimental results to start from.
A natural starting point for an analysis of differential performances obtained through the use of Boolean network models is that concerning the dynamical regimes in which they operate.
So, in the next sections we investigate what Boolean network dynamical regimes---ordered, critical or chaotic---provides an advantage for robots equipped with the adaptive mechanism we have just introduced.

\subsection{Experimental setting}

\begin{figure}[t]
\centering
\includegraphics[scale=0.5]{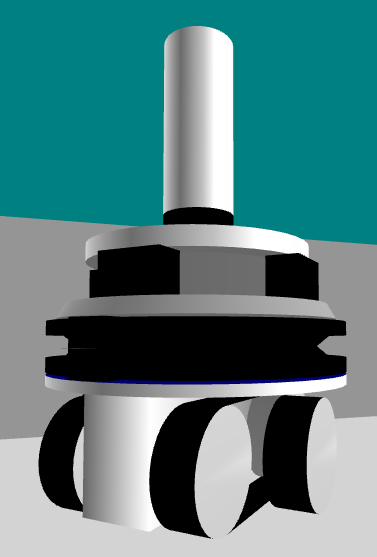}
\caption{The robot used in the experiments.}
\label{fig:footbot}
\end{figure}
\begin{figure}[t]
\centering
\includegraphics[scale=0.5]{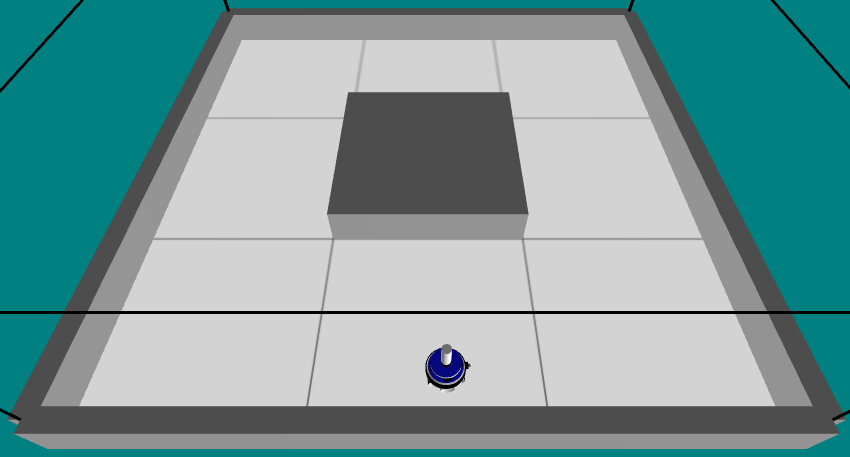}
\caption{The arena used in the experiments.}
\label{fig:arena}
\end{figure}

In our experiments we used a robot model equipped with 24 proximity sensors (evenly placed along its main circumference) and controlled by two motorised wheels (see figure~\ref{fig:footbot}). The robot moves inside a squared arena, delimited by walls, with a central box (see figure~\ref{fig:arena}). The goal we want the robot to achieve is to move as fast as possible around the central box without colliding against walls and the box itself. The robot is controlled by a BN.
The coupling between the BN and the robot is as follows: two nodes are randomly chosen and their value is taken to control the two motors, which can be either ON (node with value 1) or OFF (node with value 0) and control the wheels at constant speed. The sensor readings return a value in $[0,1]$ and so are binarised by a simple step function with threshold $\theta$\footnote{In our experiments we set $\theta = 0.1$}: if the sensor value is greater than $\theta$, then the BN node is set to 1, otherwise it is set to 0.  The 24 sensors are randomly associated to 24 randomly chosen nodes in the network, excluding the output ones. At each network update, the binarised values from the sensors are overridden to the current values of the corresponding nodes, so as to provide an external signal to the BN.

The adaptive mechanism consists in randomly rewiring $q$ connections between sensors and BN nodes (excluding output nodes, of course). The actual value of $q$ is randomly chosen at each iteration in $\{1,2,\ldots,6\}$. The robot is then run for $T=1200$ steps (corresponding to $120$ seconds of real time); if the current binding enables the robot to perform better, then it is kept, otherwise it is rejected and the previous one is taken as the basis for a new perturbation. We remark that the binding between proximity sensors and BN ``input'' nodes is the only change made to the network: in this way we address the question as to what extent a random BN can indeed provide a sufficient bouquet of behaviours to enable a robot to adapt to a given (minimally cognitive) task. 

BNs are generated with $n$ nodes, $k=3$ inputs per node and random Boolean functions defined by means of the bias $b$, i.e. $b$ is the probability of assigning a 1 a truth table entry. In the experiments we tested $n \in \{100,1000\}$ and $b \in \{0.1,0.21,0.5,0.79,0.9\}$. According to~\cite{sole_critical_points}, random BNs with $k=3$ generated with bias equal to 0.1 or 0.9 are likely to be ordered, with bias equal to 0.5 are likely to be chaotic and bias equal to 0.21 and 0.79 characterises criticality.\footnote{Along the critical line, $k$ and $b$ are linked by this relation: $k = \dfrac{1}{2b(1-b)}$.} Only the BN nodes controlling the wheels have function randomly chosen always with bias $0.5$; this is to avoid naively conditioning the behaviour of the robot, which would tend to be always moving (resp. resting) for high biases (resp. low biases). This choice has anyway a negligible contribution to the overall dynamical regime of the network. 

The performance is evaluated by an objective function that is accumulated along the robot execution steps and then normalised. The function is defined as follows:

\begin{center}
\begin{math}
 F = (1-p_{max}) \; (1-|v_l-v_r|) \; \frac{(v_l+v_r)}{2}
\end{math}
\end{center}

\noindent
where $p_{max}$ is the maximal value returned among the proximity sensors, and $v_l$ and $v_r$ are the binarised values used to control the left and right motor, respectively. The intuition of the function is to favour fast and as much straight as possible trajectories far from the obstacles~\cite{nolfi-evorobot-book}. 
Experiments are run in simulations with ARGoS~\cite{pinciroli2012-argos}.\footnote{The controller has been implemented in Lua and it is available from the authors upon request; raw data of the results are available as well.}

\subsection{Results}
\begin{figure}[t]
\centering
\includegraphics[scale=0.38]{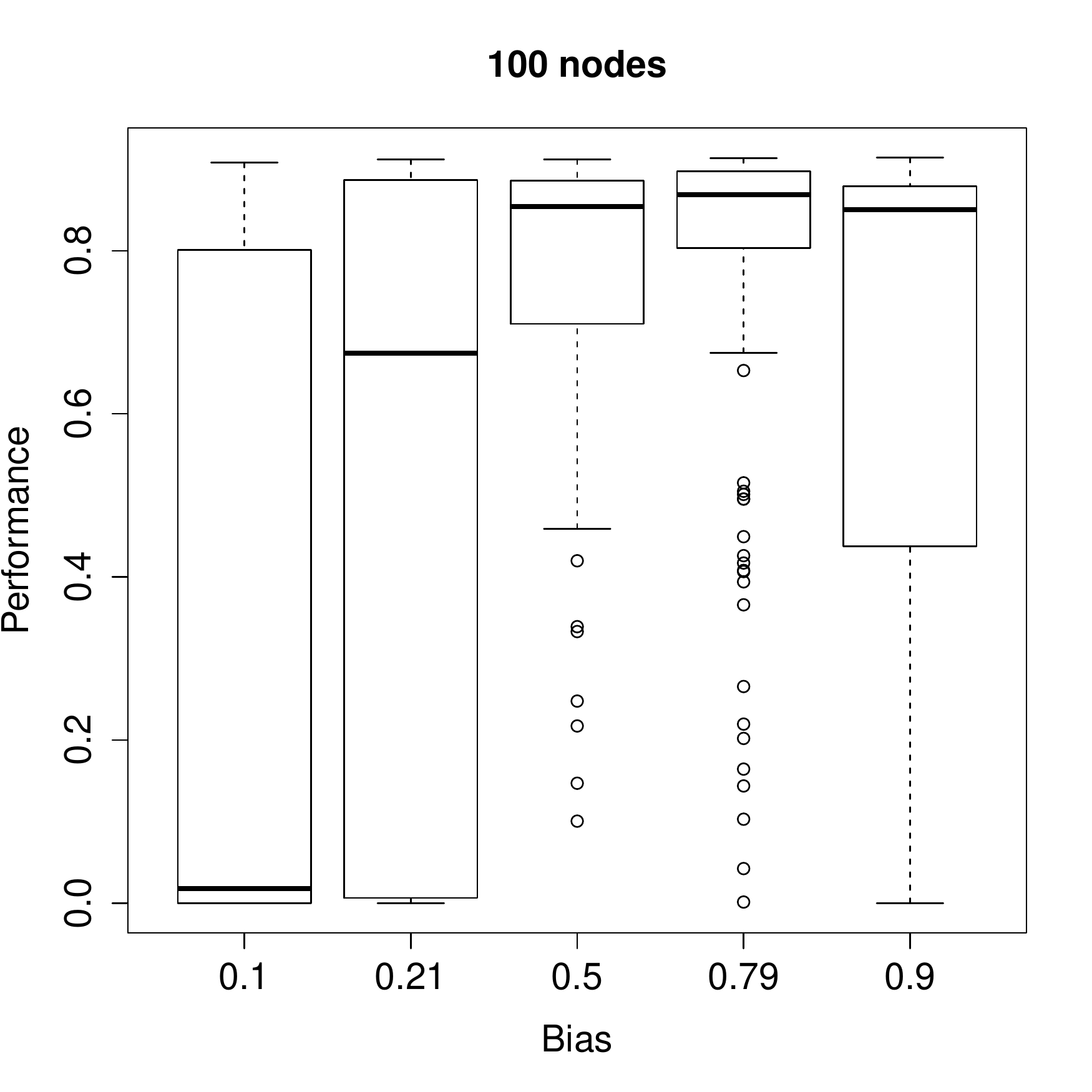}
\includegraphics[scale=0.38]{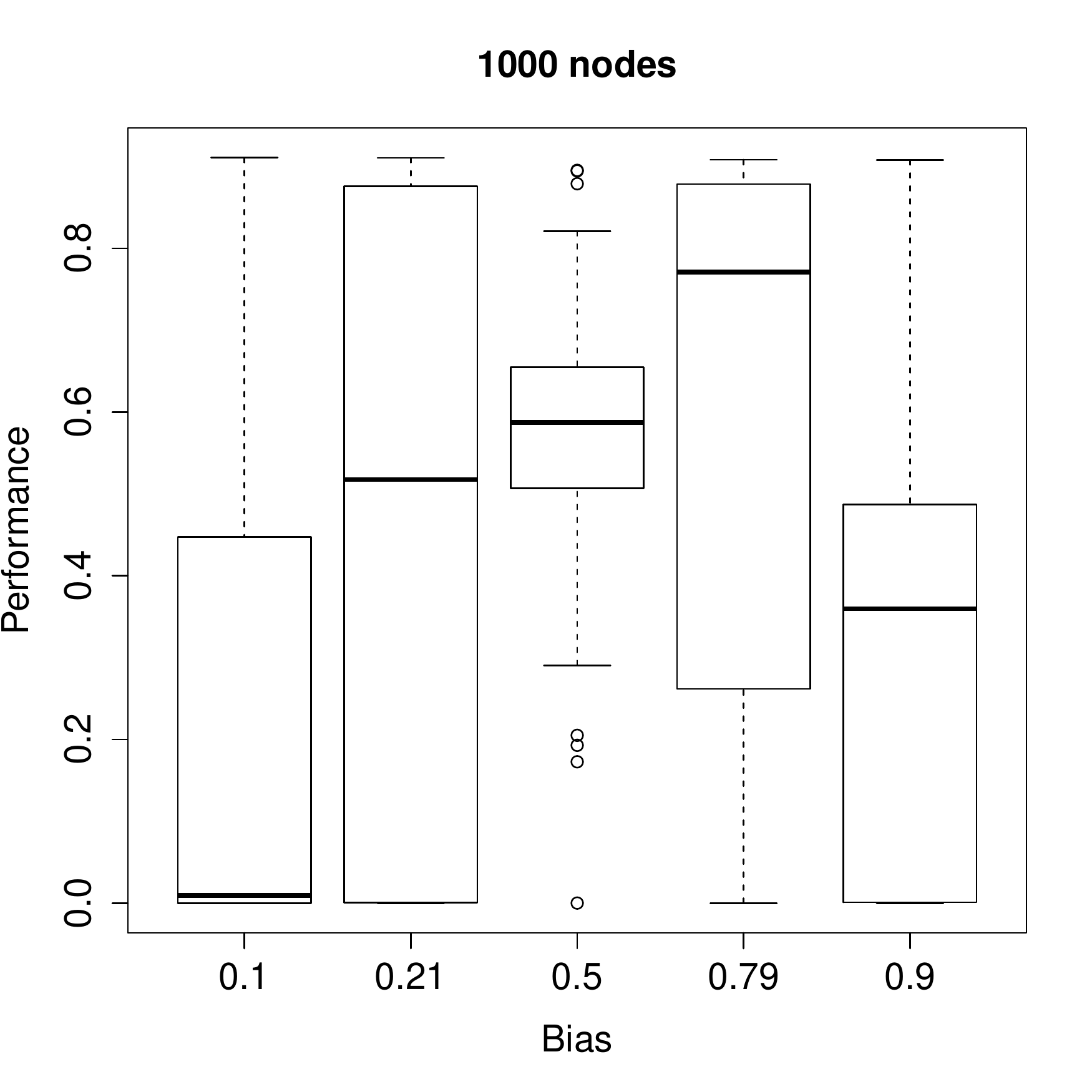}
\caption{Boxplots summarising the performance as a function of BN bias for BNs with $n=100$ and $n=1000$.}
\label{fig:xor}
\end{figure}
\begin{figure}[t]
\centering
\includegraphics[scale=0.38]{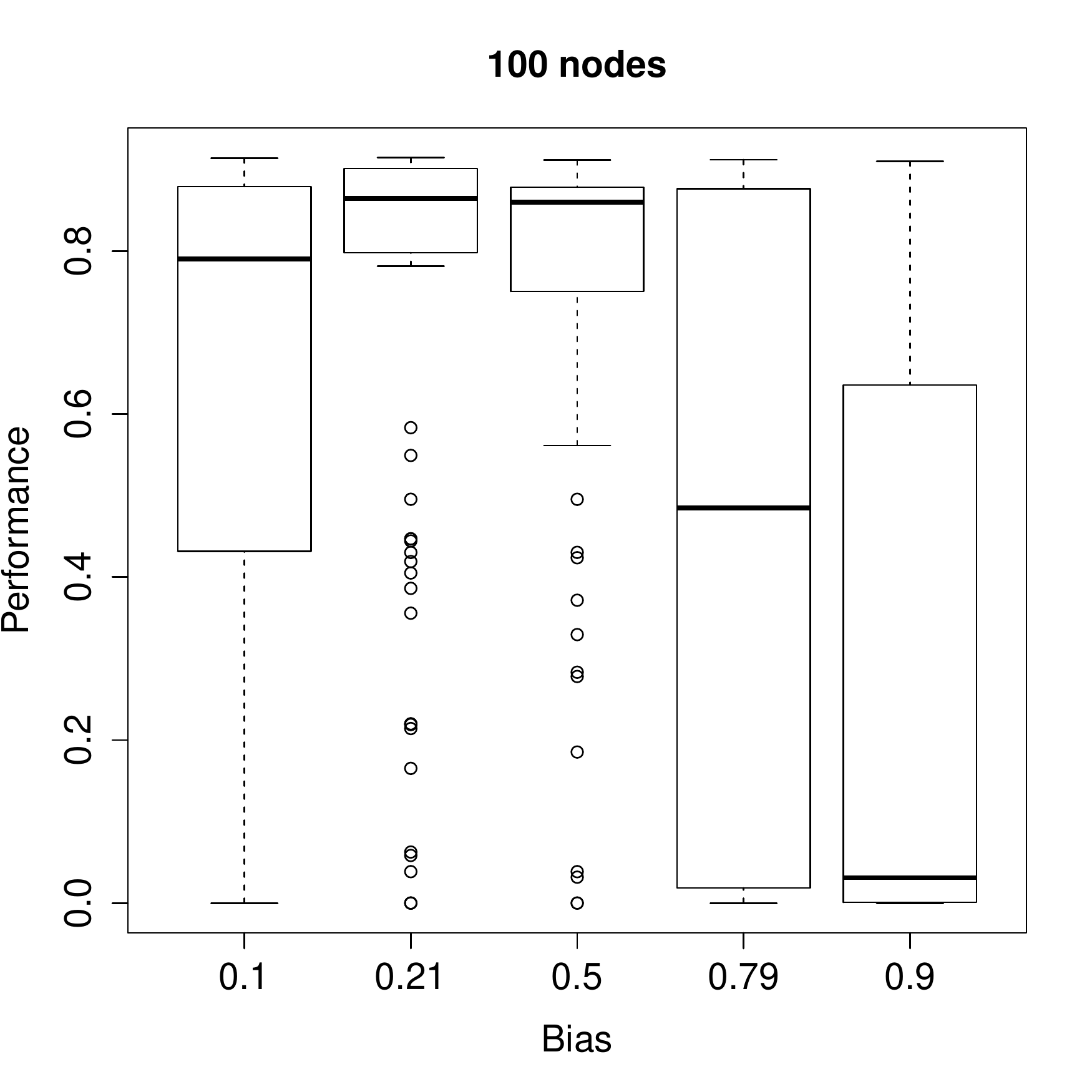}
\includegraphics[scale=0.38]{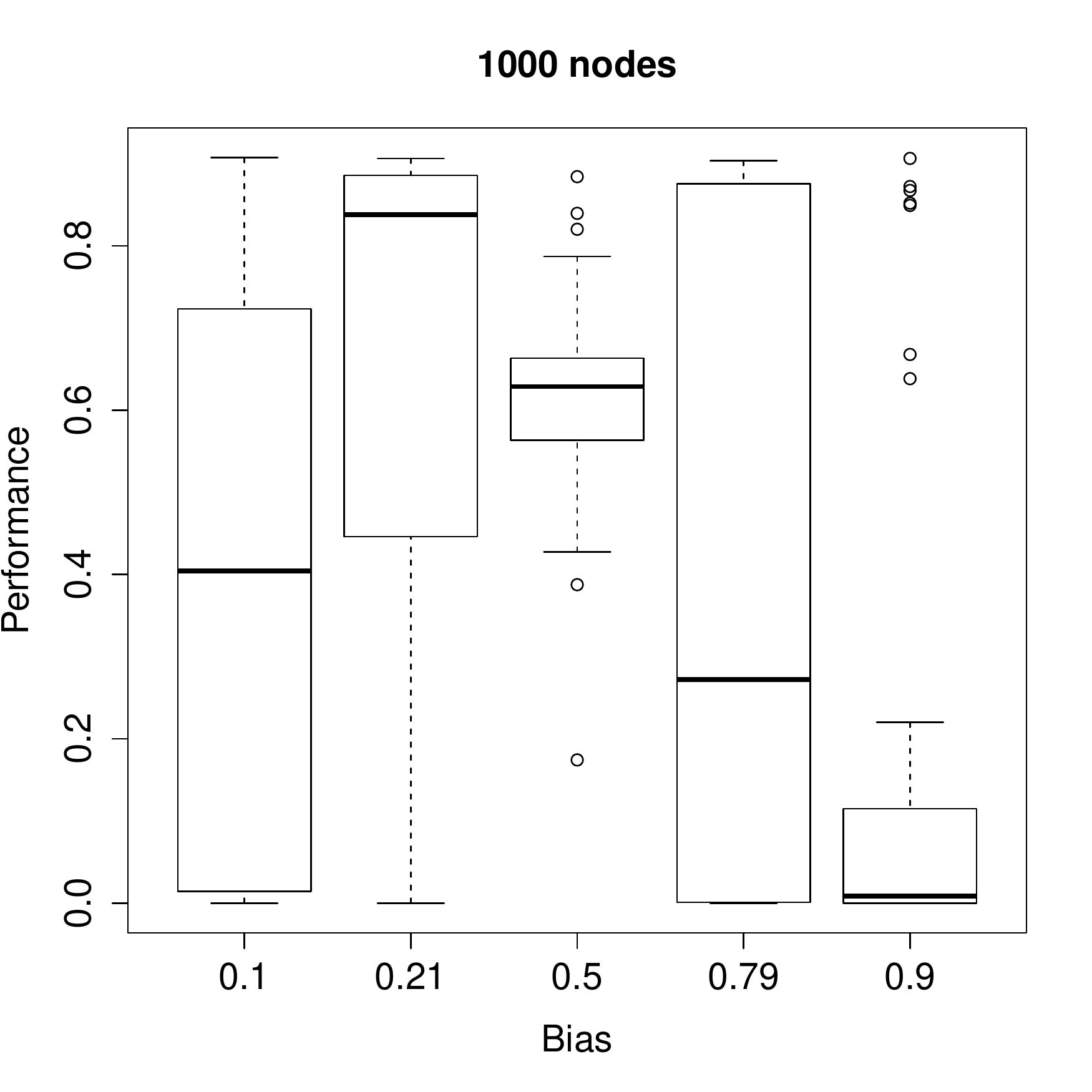}
\caption{Boxplots summarising the performance as a function of BN bias for BNs with $n=100$  and $n=1000$ for robots controlled by BNs with a dual encoding (i.e., 0 denotes that an obstacle is detected).}
\label{fig:xor-dual}
\end{figure}

We run 1000 random replicas for each configuration of BN parameters and collected statistics on the best performance attained after a trial of $1.44 \times 10^4$ seconds (corresponding to $1.44 \times 10^5$ steps in total). In order to avoid variance due to the initialisation of the network, all nodes are initially set to 0.
Since the evaluation function can not be maximal across the whole run, as the robot must anyway turn to remain in the arena and avoid the obstacles, values of $F$ greater than 0.7 correspond to a good performance.
As we can observe in figure~\ref{fig:xor}, despite the simple adaptation mechanism, a large fraction of BN attains a good performance.  Notably, critical networks attains the best performance---this result is striking for large BNs ($n=1000$). 
In particular, for $n=100$ the results achieved with $b=0.21$ are significantly better (Wilcoxon test, with $\alpha=0.05$) than all the other cases, with the exception of $b=0.5$ for which we could not reject the null hypothesis. As for $n=1000$ the case with $b=0.21$ is significantly better than all the other ones.
We observe, however, that just one of the two bias values corresponding to the critical regime corresponds to a good performance. The reason is that in our experiment the symmetry between 0 and 1 is broken, because a 1 means that an obstacle is detected. To test the robot in the dual condition, we ran the same experiments with a negative convention on the values (if the obstacle is near, then the node is set to 0; similarly, the wheels are activated if the corresponding output node state is 0\footnote{We kept this dual condition uniformly across all the choices, even if, being the bias of output nodes 0.5, the encoding has no effect on average on the wheels.}). As expected, results (see figure~\ref{fig:xor-dual}) are perfectly specular to the previous ones (and the same results of the statistical test hold).

\section{Discussion}\label{discussion}

The picture emerging from our experiments is neat: one bias value characterises the best overall performance and this value is one of the two along the critical line. The reason of the asymmetry between the two critical bias values has to be ascribed to the symmetry breaking introduced by the binarisation of the sensor values. Anyway, the remarkable observation is that random BNs generated with a bias corresponding to critical regime adapt better than the other kinds of BNs. Since the adaptive mechanism only acts on the mapping between sensors and input nodes, the dynamical regime of the BNs is preserved; therefore, we have a further evidence that critical BNs achieve the best performance in discriminating the external signals.

One might ask as to what extent the adaptive mechanism we have implemented can be said to be a case of phenotypic plasticity. To answering this question we first observe that, in our setting, adaptation involves only the way external information is filtered by the robot and so it concerns the sensing module of the system; second, this adaptation takes place during the ``life'' of the individual and it is based on a feedback that rewards specific behaviours (i.e. those favouring wandering with collision avoidance), without changing the actual ``genetic'' structure. In other words, our mechanism mimics a kind of sensory development tailored to the specific environment: in general, the robot can be coupled with the environment in a huge number of possible combinations, each constraining the system to express a particular behaviour (phenotype); the mapping between sensors readings and network nodes is the result of the embodied adaptation of the sensory-motor loop and manifests one particular phenotype, emerged from the interaction between the robot and the environment.

\section{Conclusion}\label{conclusions}

In this work we have shown that robots controlled by critical BNs subject to sensor adaptation achieve the highest level of performance in wandering behaviour with collision avoidance. The sensor adaptation mechanism used in this work consists in varying the coupling between robot sensors and BN nodes---whose value is then set by the binarised reading on the associated sensor. Other possible adaptive mechanisms can be chosen, e.g. varying the coupling between BN nodes and robot actuators, and can also be combined; in general, structural modifications of the BN are also possible, such as the ones acting on Boolean function, network topology and also network size. These adaptive mechanisms are subject of ongoing work and preliminary results suggest that sensor and actuator adaptation mechanisms are way better than structural ones, and that critical BNs again attain superior performance.
As a next step, we plan to investigate the relation between criticality of controlling BNs, their performance and the maximisation of some information theory measures, such as predictive information~\cite{ay2008predictive}, integrated information~\cite{edlund2011integrated} and transfer entropy~\cite{lizier2008information}.

Besides providing evidence to the \textit{criticality hypothesis}, the results we have presented make it possible to speculate further: criticality may be a property that enables phenotypic plasticity---at least as long as sensory adaptation is concerned. We believe that this outcome provides a motivation for deeper investigations, which may be  primarily conducted in simulation or anyway with artificial systems. Nevertheless, we also envisage the possibility of devising wet experiments, in which the dynamical regime of an organism is externally controlled and its ability to exhibit phenotypic plasticity can be estimated.

In addition, a mechanism like the one we have introduced may be an effective tool for tuning artificial systems to the specific environment in which they have to operate. As a futuristic application, we imagine the construction of miniaturised robot that can accomplish missions precluded to humans, such as being inoculated into higher organisms to repair them, or recovering polluted environments. 
In fact, recent technological advances have made it possible to build  incredibly small robots, till the size of tens of nanometers. The current smallest robots---built by biological matter---can perform only a few predetermined actions,\footnote{See, e.g. the recent prominent case of Xenobots~\cite{xenobots}} therefore they can not attain the level of adaptivity and robustness needed for a complex mission. On the other hand, Artificial Intelligence software has recently made tremendous advancements and has been proved capable of learning and accomplishing difficult tasks with a high degree of reliability. This software, however, can not be run onto tiny robots. A viable way for filling this gap is provided by control programs based on unconventional computation, such as the ones derived from cell dynamics models, where phenotypic plasticity may play an important role.

\end{document}